# A Cost-Efficient Approach for Creating Virtual Fitting Room using Generative Adversarial Networks (GANs)

Kirolos Attallah[1], Girgis Zaky[2], Nourhan Abdelrhim[3], Kyrillos Botros[4], Amjad Dife[5], Nermin Negied[6]

Faculty of Electrical and Computer Engineering, University of Ottawa, Ottawa, Canada[1, 2, 3, 4, 5]
School of Information Technology and Computer Science, Nile University, Giza, Egypt[6]

*Abstract*—Customers all over the world want to see how the clothes fit them or not before purchasing. Therefore, customers by nature prefer brick-and-mortar clothes shopping so they can try on products before purchasing them. But after the Pandemic of COVID19 many sellers either shifted to online shopping or closed their fitting rooms which made the shopping process hesitant and doubtful. The fact that the clothes may not be suitable for their buyers after purchase led us to think about using new AI technologies to create an online platform or a virtual fitting room (VFR) in the form of a mobile application and a deployed model using a webpage that can be embedded later to any online store where they can try on any number of cloth items without physically trying them. Besides, it will save much searching time for their needs. Furthermore, it will reduce the crowding and headache in the physical shops by applying the same technology using a special type of mirror that will enable customers to try on faster. On the other hand, from business owners' perspective, this project will highly increase their online sales, besides, it will save the quality of the products by avoiding physical trials issues. The main approach used in this work is applying Generative Adversarial Networks (GANs) combined with image processing techniques to generate one output image from two input images which are the person image and the cloth image. This work achieved results that outperformed the state-of-the-art approaches found in literature.

## I. INTRODUCTION

Virtual fitting rooms (VFR) bring great opportunities to the fashion industry by enabling consumers to virtually try on products. However, while VFRs have technically been available for a while, they are less utilized because of many reasons, amongst them the consumers' potential concerns of accuracy of the simulation, the cost of the technologies used in building VFRs like Kinect cameras and depth sensors, and the difficulty of using them by the customer because of the special settings they require like special types of LCDs and mirrors. Research has proven that online clothes purchases had increased, and the return requests had decreased after the intervention of virtual fitting rooms (VFRs) [1]. e-Commerce development, AI development, and pandemics shared in both the research and industry demand of VFRs. In this paper two novel outputs are obtained and explained, the first one is a new dataset which is large and various (i.e., containing images for males and females in different poses and from different fashion houses, without excluding any challenges from the collected images) for the purpose of building VFRs, and the second one is a real time, cost efficient, portable, easy to use, and accurate VFR. The rest of this paper is organized as follows: Section II reviews the work done in literature to address this problem. Section III explains the methodologies and techniques used in this work to build the VFR. Section IV demonstrates the experiments conducted to evaluate the work. Section V discusses the results. Finally, the paper is concluded in Section VI.

## II. LITERATURE REVIEW

Recently, the idea of virtual fitting rooms has attracted researchers because of the emergence and development of virtual and augmented reality. The digital revolution and affordable technologies and devices also made virtual fitting rooms an area of interest. According to literature there are eight different types of virtual fitting rooms (VFR), which are: Body scanning VFRs, 3D avatar VFRs, 3D customer's modelling VFR, 3D mannequin VFR, Augmented reality VFRs, Robotic mannequins, Dress-up mannequins for mix-and-match, and the real fashion model VFR [1 & 2]. From all technologies and types of body scanning machines proved to be the most accurate, but the most expensive at the same time [3-6]. Although the large variety of technologies that could share in maximizing the customer experience while trying the cloth item online, the cost plays a major role of the types existing in the market. Some features and complementary methods could share in maximizing the customer experience besides the VFRs such as: fit guides, size charts, comparison avatars, virtual cat walks, brands comparisons, etc. [7 - 8].

Researchers used many emergent technologies to simulate fitting rooms, like augmented reality (AR), virtual reality (VR), depth information using Kinect cameras, deep learning approaches (DL), 3D reconstruction, and hybrid approaches. Pereira et al. in 2010 [9], used AR and Kinect camera to establish virtual fitting room using depth information, the authors tested their approach using Open CV and Open GL techniques with different six degrees of human head detection, but they mentioned nothing factual about their results.

Kostas [10] used the same technologies (Kinect cameras and AR) to implement virtual fitting room for trying different cloth items, but the higher accuracy they reached was 67%. Dias et al. [11] also used the same approaches for the same purpose, but they added different features, and they developed a better user interface, but they mentioned nothing about the VFR results. Below is a sample of their results. Hashmi et al. [12] used the AR combined with Haar-cascades classifier to implement VFR. The authors tested their classifier using 50 subjects with 10 different dresses and they confirmed that their approach had outperformed other approaches in literature. Mehta et al. [13] had used AR, VR, and Mixed reality (MR) combined with Head Mounted Displays (HMD). The authors claimed that their VFR can give customers better online shopping experience, but again they mentioned nothing factual about the results.

Boonbrahm et al. [14] in 2015 used VR to build VFR. The authors considered the differences between materials of clothes, and they confirmed that this matters in the final appearance of the dressed person. The materials they considered were jean, satin, silk, and cotton. França & Soares [15] in 2018 have discussed the idea of a complete simulation of Virtual Fitting Room (VFR) using virtual reality (VR) in which the customer feels he/she is existing in a real fitting room. Alfredo et al. [16] used Kinect camera along with gesture recognition and cloth transfer algorithm to create a virtual try in application. Despite the authors concluded that their approach combined with the depth information obtained using the Kinect camera made the customer experience more enjoyable, they mentioned nothing factual about the results.

Sapio et al. [17] in 2018 integrated different body scanning technologies to create a VFR. The authors used Kinect combined with other more expensive body scanners, but they found out that their approach is not only expensive but also needs human intervention and cannot be considered as an easy automated solution.

Silvestro et al. [18] in 2020, used avatars to allow the user to select the size and the shape, but regarding their results the authors mentioned that they are still working on their project to get good results. Ileperuma et al. [19] in 2020 used the CNNs combined with augmented reality (AR) to detect human body and create an image for a dressed object. The authors claimed to achieve 99% accuracy for their generated images. Nande et al. [20] in 2021 used generative adversarial networks (GANs) to replace the cloth item the customer already wearing, by the cloth item the customer wants to buy in a new image. The authors confirmed that they achieved structural similarity index measure (SSIM) matrices of 0.8.

Chen et al. [21] in 2021 also used M5 transformer to build the VFR and they compared their results to other state-of-the-art DL approaches in literature and they achieved very good results. Lyu et al. [22] in the same year, the authors proposed a High-Resolution Virtual Try-On network (HR-VTON) model to synthesize virtual fitting images, which consists of three sub-modules, namely, a clothing matching module, a try-on module and a refine module. They tested their proposed model on Zalando datset and they confirmed that they achieved accuracy rate of 81%. In 2021 Hyder et al. [23] studied the capabilities of Microsoft Kinect sensor and the role of augmented reality in simulating the surrounding world. The authors confirmed that 85% of the volunteers experienced their system and recommended it as a good 3D learning system. Singh et al. [24] in 2021 also, used Generative Adversarial Networks (GANs) to create a VFR. The authors implemented three models which are: the semantic generation module, the clothes wrapping module, and the content fusion module, and they tested their approach using the Zolando dataset also, and they confirmed that their approach achieved very robust results.

In fact, Lye et al. and Singh et al proved to obtain the best results in literature, but on limited dataset with very narrow variety of images as shown in Section IV(A)(1). Chandani et al. [25] in 2022 tried many methods to build a VFR like AR, ANNs, and CNNs. The authors claimed that the users can move freely infront of camera, and change the colors of the outfit, but they mentioned nothing numeric about the mean error or the accuracy rate. Malathi et al. [26] in 2022 also used DL to reconstruct the 3D graphical perception of the user from his/her 2D image to create cost efficient VFR rather than using Kinect camera. The authors confirmed that their approach achieved very good results, with many useful features that maximize user experience satisfaction.

Prabhakar [27] et al. in the same year, used Alpha Channel Masking to mask the user's shirt to gain an area of interest. The authors only worked on three different colours of the same style (Shirt). Mohamed et al. [28] in 2023 used the 3D reconstruction of human body to build VFR. The authors used Kinect combined with other more expensive body scanners, but they found out that their approach is not only expensive but also needs human intervention and cannot be considered as an easy automated solution. The authors used neural networks to reconstruct the 3D models of the human body and the cloth item and put them together in one photo. The authors also used CNNs combined with ResNet-50 to identify the type of texture to make the reconstructed image more accurate and closer to reality, but in their results, they showed a bit large mean error.

Omkar et al. [29] in 2023 have combined AR with DL to build a VFR. The authors confirmed that their approach that renders the image cloth on the customer's image improved the customer's experience, but they also mentioned nothing factual about their results. Yang et al. [30] in 2023 considered the body mass (BM) of the user and they confirmed that their approach improved the virtual fitting results to a great extent.

## III. METHOD

This work is designed to be deployed on a mobile application in which the process would be much easier for the consumer to do and much cheaper for the clothing shops. The proposed system mainly consists of three main phases: image preprocessing, the GANs part consists of a conditional generator to generate a new segmentation map with the person wearing the new cloth item, and the image generator to generate the final image. The generic flow of the system is as follows: The system asks the user to enter two input images which are, the person image in any pose, and an image for the cloth item he/she wants to buy, to generate a new image of the person wearing the new cloth item. Following is the explanation of every module. Fig. 1 demonstrates the complete architecture of the proposed system.

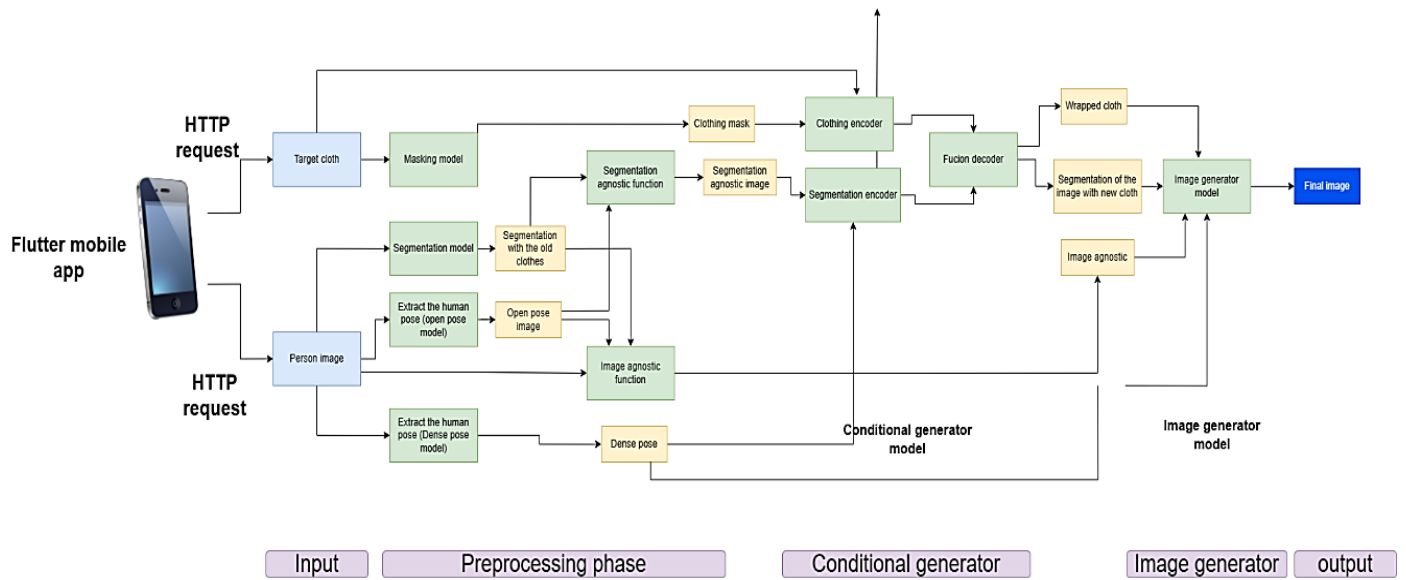

Fig. 1. The complete architecture of the proposed system.

### A. Images Preprocessing

The input image preprocessing module is composed of four different components: OpenPose, DensePose, segmentation, and cloth mask (see Fig. 1). As mentioned before the user should provide the system by two images, which are his/her image, and the cloth item he/she wants to buy. Regarding the cloth image, it should be segmented using a masking model to be able to recognize the cloth and separate it from the background. Regarding the person's image, some steps should be held such as: background removal, segmentation of different human body parts, definition of OpenPose key points to wrap the cloth over the person in any position not only from the front, 3D human image construction using dense-pose to enhance the final wrapped cloth on the person pose, and finally the combination of the open-pose and the dense-pose to create an agnostic image that would make the system able to focus on the key parts of the human body that should place the new t-shirt on. The outputs of the preprocessing module would be the dense-pose, clothing mask, and segmented agnostic image, which will be the inputs to the conditional generator that generates the new image. Following is a detailed explanation of the preprocessing steps.

*1) Human body segmentation:* Human parts segmentation is the most important preprocessing step for the generator, as the following modules of the proposed system depend on its results. There are 20 parts in the human body image that should be segmented and classified accurately to identify the clothes' location and these classes are: Background, hat, hair, gloves, sunglasses, upper clothes, dress, coat, socks, pants, torso skin, scarf, skirt, face, eft arm, right arm, left leg, right leg, left shoe, and right shoe (see Fig. 2).

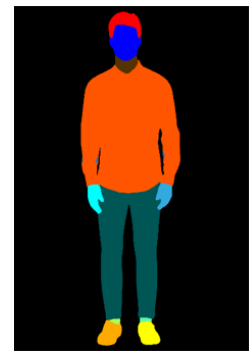

Fig. 2. Typical human body segmentation sample.

The Crowd Instance-level Human Parsing (CIHP) model [31] was used to segment the parts of the human body. This model relies on the concept of part grouping network (PGN) [32] where the input image is scaled into six scales which are: 0.5, 0.75, 1, 1.25, 1.5, 1.75 of the original width and height. The input to this model should include the human image its invert. In other words, the input to this model is a stack of image pairs; the image and its invert [image, image invert]. For each input the average of scales is calculated for every input image. This pipeline consists of six models makes the inference time very large (140 sec), but it gives very accurate results, however, this huge inference time is not suitable for this use-case, so we thought how to decrease it. To solve the problem of large inference time we tried other models for segmentation such as self-correction human parsing [33], and Deep-Lab V3 [34]. After utilizing and evaluating the three models Deep-Lab V3 model were selected for this work (see Section V).

*2) Agnostic images generation:* Agnostic images are created to eliminate all the old clothes from the input image and the segmented image. First, a color scale conversion from RGB to gray is done for both input and segmentation images. Then, the background color is converted to black.

This can be done using the open pose key points and the segmentation colors map which help us to define where the upper parts are to eliminate them, and we used the key points to mask the hands only because the segmentation gives the information of where the left and right arms are. Optimization was done using the high-quality segmentation and open-pose key points. Fig. 3 represents the agnostic image generation steps.

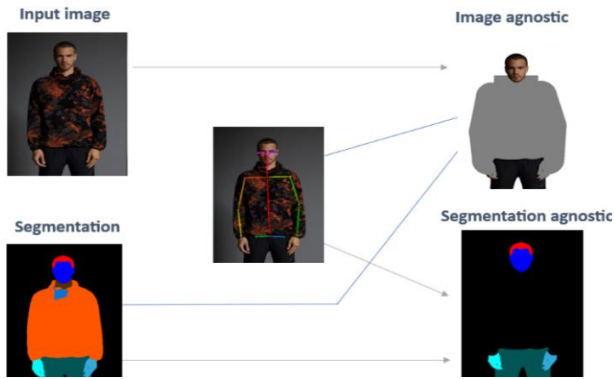

Fig. 3.    Agnostic images generation.

After many experiments, we found that eliminating the background gives us more accurate results because the dataset distribution which the model trained on with a white background, The background class is one of the segmentation classes, so we exploit that to eliminate the background by looping on the image and change the color of it to white using open cv (see Fig. 4).

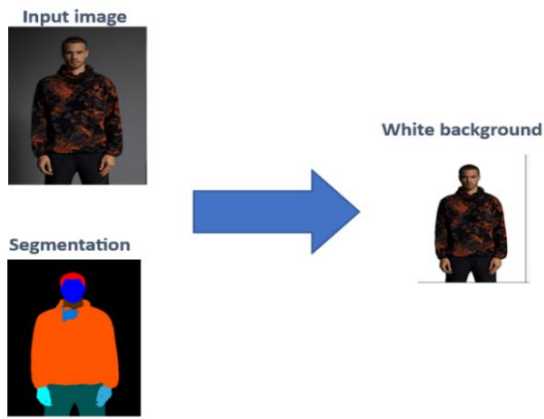

Fig. 4.    Background removal example.

*3) Cloth masking:* Cloth masking is one of the main modules of the proposed system. To build a model that can generate the mask of the clothes; two approaches were suggested in this work, in the first approach, three models were evaluated; AIM, Timi-Net, and P3M model to detect the main object in the image and remove other objects, so, they can be used to generate the mask of the clothes if the clothes considered as the main object in the image. The second approach is cloth segmentation which is used to generate cloth mask in two steps instead of one step first by performing segmentation and then binarizing the result. After evaluation and analysis of the suggested models, the cloth-segmentation was found to be the champion model which is a robust model against the different colors, and backgrounds [35 - 37]. Fig. 5 shows a sample result for cloth masking.

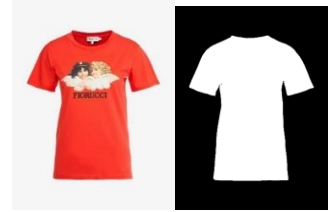

Fig. 5.    Cloth masking sample result.

*4) Using dense pose:* The fourth and final sub-module in the preprocessing module is the DensePose, in which (R_101_FPN_DL_s1x) [38] was used and succeeded in achieving the best result compared to the baseline model used by Lyu et al [22] (see Section V). Fig. 6 shows the difference between the DensePose proposed by this work, and the DensePose of the work in literature.

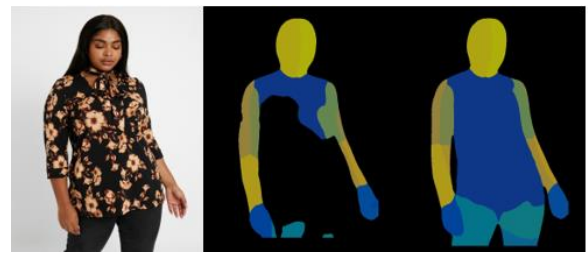

Fig. 6.    Enhanced DensePose (right) vs. original or literature's DensePose (middle).

### B. Output Image Generation using Conditional GANs

The second phase of the system is to generate a new image for the customer wearing the new cloth item, GANs was chosen and deployed because of its ability to generate new images with high accuracy. The GANs use a conditional generator where a wrapped cloth and a segmentation map of the person wearing the new cloth are the targets. There are two encoders with five residual blocks; cloth encoder and segmentation encoder, and one decoder that take these targets as conditions to get the most accurate cloth image appearance flow and segmentation map. Then the next fusion block in the encoder takes the previous output after passing to a 3x3 convolutional layer to predict the appearance flow-map from the flow pathway and segmentation features from the seg pathway. Information is exchanged through these two pathways to get the most accurate segmentation map and appearance flow. The last block of the decoder is the conditional aligning that removes the overlapping regions from the cloth mask (convert the straight mask to be fitted on the person image) and handle occlusions (remove hands or any parts from person image in front of the cloth). The following hyperparameters were used: pixel wise cross entropy loss function, Perceptual and L1 losses for best wrapping of the clothes, in addition to least-squared GAN loss. Also, multi-scale discriminators were applied for the conditional adversarial loss calculation.

### C. Image Generator

This phase includes a series of residual blocks, up sampling layers, Spade normalization layers based on the segmentation map parameters. It takes the image agnostic, new segmentation map, dense pose and wrapped close resized and concatenated to the activation. To evaluate the generated image a discriminator rejection was deployed to filter segmentation maps with low quality. It is based on data distribution and implicit distribution from the generator. If the output image is of very low quality, it would be rejected.

## IV. EXPERIMENTAL WORK

We have tested some architectures in literature, and it is found to be working perfectly on females as the authors selected only the females' images from Zolando Dataset, however, accordingly, their approach didn't work with any image outside their dataset as shown in Fig. 7. This fact tells that there is an overfitting in the results found in literature. To avoid the overfitting problem in this work, new images from multiple websites were scrapped and trained beside the dataset found in literature. The following section describes both the dataset in literature and the dataset collected for this work.

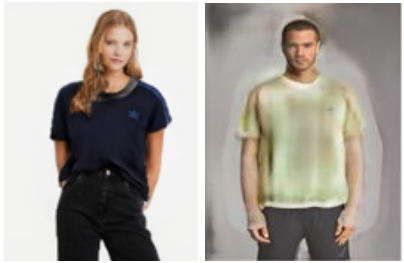

Fig. 7. Results obtained from their dataset (top) vs. results obtained using image outside the dataset (down).

### A. Datasets

There are two different datasets used in this work for training and validation, one of them is open-source data used in literature and the other one is collected for this work. The following sub-sections describe the datasets used in this study.

*1) Zolando dataset:* Zolando is a high-resolution dataset that is used for virtual try-on tasks. It consists of the frontal view woman ad top clothing image pairs. It is split into about 11K training pairs and 2K testing pairs. This dataset was used by Iye et al. [22], who obtained the best results in literature, and for that reason we used their approach as a baseline model to compare the proposed approach with.

*2) New dataset collected in this work:* About 5K images of males' images from Zara, Farfetch, and Zalando websites were scrapped to avoid the overfitting issue. The collected dataset is publicly available on Kaggle [39]. Some challenges in the collected dataset are tattoos, birth marks, caps, dark sunglasses, beard, head cuts, Black and Asian people, and complicated backgrounds (see Fig. 8). The dataset collected was meant to cover a diverse dataset distribution, and balance between males and females' images as follows: images from Zolando dataset (females), images from Zolando (males), images from Zara (males), images from Farech (males), and images from Farech (females).

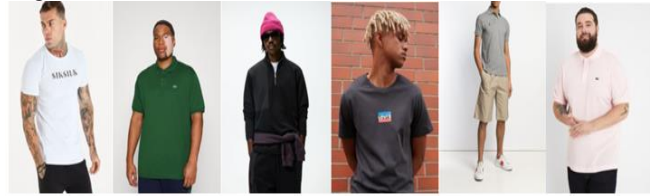

Fig. 8. Some challenges in the new scrapped dataset collected in this work.

### B. Validation and Verification

The system was validated at the submodules level starting from the cloth masking, going through DensePose, OpenPose, segmentation, closing mask, the application itself, and reaching the deployment of the app on AWS server. The integrated submodules were also tested using many test cases and the proposed approach perfectly outperformed the state-of-the-art approaches. Based on the new implementation of the segmentation mask, the response time was reduced from 4 minutes to 78 seconds. The model was deployed on AWS successfully and the application received requests and responded concurrently. Fig. 9 shows the difference between our output (right) and the state-of-the-art result (middle) for the same image. Fig. 10 shows the effect of clothes masking and DensePose, in which it can be noticed that there is a great enhancement of the clothes' alignment.

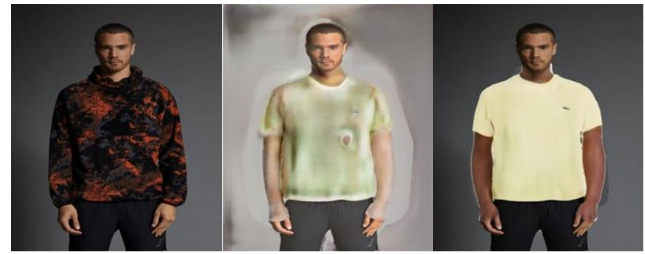

Fig. 9. The difference between the proposed work output (right) and the state-of-the-art result (middle) for the same images.

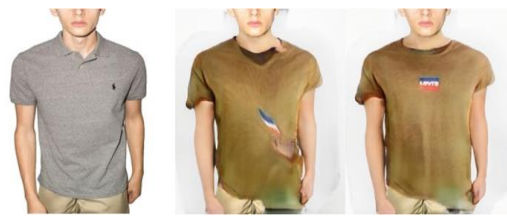

Fig. 10. The state-of-the-art results (middle) vs. the proposed approach results (right).

## V. RESULTS AND DISCUSSION

This section demonstrates the results achieved by this proposed work and discusses them. The following subsections show every submodule, the results, and the analysis of these results.

### A. Segmentation

CHIP, Self-correction human parsing, and Deep-Lab V3 segmentation models were used and compared in this work to select the best as mentioned before. The comparison was done in terms of segmentation results and execution time. The following sub-section describes every model and the findings of using it.

*1) Self-correction human parsing segmentation:* To increase the dependability of both the learned models and true labels, the self-correction model uses a learning scheduler to infer more trustworthy pseudo-masks by repeatedly aggregating the learned model with the former ideal one in an online learning. Using an annotation initialization model as the first step in the learning process, and then adjusting the labels according to the data can improve the model's performance. Through cycles of self-correction learning, both the models and labels improve in accuracy and strength. This model has a good inference time (2 sec), but its pre-trained weights were on a different dataset which missing the torso skin class (the neck class). The pre-trained weights were on the Look into person (LIP) dataset [40] which contains the same classes, but the only difference is the torso-skin class is replaced with a jumpsuit, so the results were without the torso-skin class. To address the problem of the missing torso class, the model was trained on another dataset including all classes, but the results were worse than the CHIP model, and it required more computational resources for training and testing (see Fig. 11).

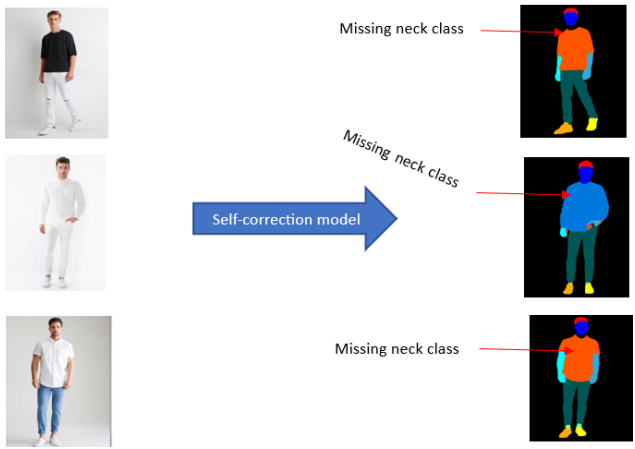

Fig. 11. Sample of self-correction segmentation results.

*2) Deep-Lab V3 model segmentation:* DeepLab V3 utilizes a novel architecture that combines multi-scale features, dilated convolutions, and conditional random fields to produce high-quality segmentation masks. It is highly effective at handling complex scenes, such as those found in urban environments, and achieves superior performance compared to other state-of-the-art segmentation models. The model was trained to segment all parts within much less time. DeepLab V3 model worked in 0.23 seconds only to produce segmented image. Although the mean Intersection Over Union (IOU) is less than mean IOU of CHIP model, we have chosen DeepLab V3 over CHIP model due to the great difference in execution time, meanwhile, the mean IOU difference between both models is not so large to consider over the time. Table I demonstrates the results and execution time of the three models. Fig. 12 shows the segmentation results of CHIP and DeepLab V3 models.

TABLE I.    SEGMENTATION RESULTS AND EXECUTION TIMES OF THE USED SEGMENTATION MODELS

| Model | Time (seconds) | IOU |
|---|---|---|
| CHIP | 38.41 | 78.21 |
| DeepLab V3 | 0.23 | 64.53 |

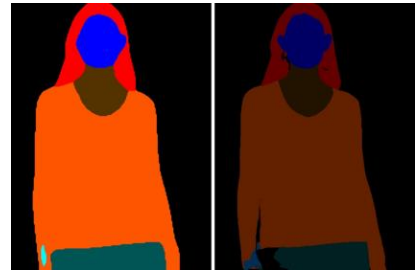

Fig. 12. Segmentation results of DeepLab V3 (right) and CHIP (left) models.

### B. Open Pose

Three different approaches were tested to find the optimal open pose model. Those three models are the Media-pipe model [41], the Detectron model [42], and the CMU-Perceptual-Computing-Lab [43 & 44]. The first model failed to address the occlusion challenge (see Fig. 13). The second model outperformed the results of the first model, but still missing some important points. The third model solved the problems of the previous two models successfully. The following figures show the differences between the results obtained by the three models. Here Fig. 14 shows Detectron open pose results and Fig. 15 shows the CMU-Perceptual open pose detection results.

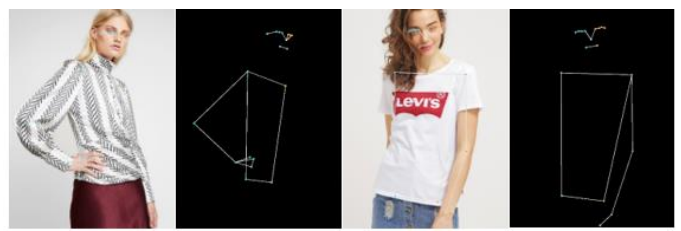

Fig. 13. Media-pipe pose detection results (occluded arms are missing).

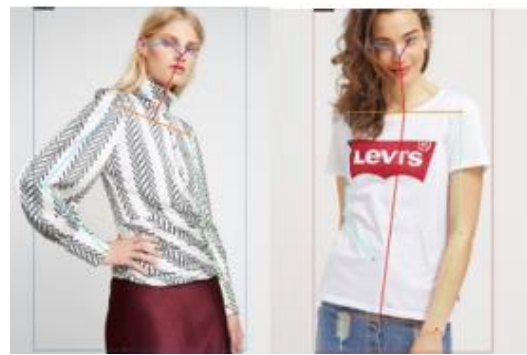

Fig. 14. Detectron open pose detection results (better but some important points are still missing).

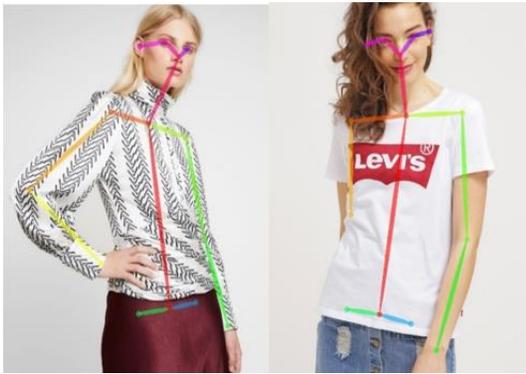

Fig. 15. CMU-Perceptual open pose detection results (the best results out of the three models).

### C. Cloth Masking

To verify the quality of the cloth masking model, three experiments have been held to check the performance of the proposed approaches to get the best cloth masking model. The first experiment was to find the best model of the three models that perform image matting, namely: AIM, Timi-Net, and P3m [45 - 48]. A dataset of 5K images dataset has been scrapped from different fashion websites, then it has been used to compare the results of each one of the three models. The AIM model obtained the worst in terms of average inference time per image, and ability to generate the mask. The abilities of P3M and Timi-Net models to generate the mask were almost the same, but P3M proved to be 1.5 faster than Timi-Net, so, the P3M model was chosen for cloth masking submodule. A manual analysis for all images was then conducted to define the strength and the failure points of the P3M model. As a result of the analysis, P3M showed great results for all cases except in the case of white clothes on white backgrounds, and to solve this problem, image binarization step was added to the segmentation module. The second experiment has been conducted to check the applicability of the second approach to generate the mask using YOLO5 and YOLO7 [49 - 53] for image segmentation, then binarizing the results. The proposed method showed high capability of detecting the existence of the clothes in the images, but at the same time YOlO5 and YOLO7 were not the best choices to perform the cloth segmentation task since they miss some of the images without detecting the existence of the clothes in the image. The third experiment has been held to check the performance of two models which are the U2Net and the cloth-segmentation model. The top 20 images that P3M models failed generating their masks were used in this experiment. U2Net failed with some white clothes with white background, on the other hand, cloth-segmentation model performs well since the result is robust against different colors and backgrounds. As a result, the cloth-segmentation model has been chosen to be integrated with the other models to complete the proposed solution.

### D. Dense Pose

Different state-of-the-art models in this area were applied, and "Detectron 2" proved to be the best amongst them. Detectron 2 was originally developed by Facebook, where 27 models were trained. All the 27 models were used and evaluated in this work to measure how suitable they are to this problem and select the best accordingly.

### E. Image Generator

This step is the final step which takes the outputs of all the steps preceding it as an input to generate the final output of the system which should be the image of the user wearing the new cloth item. Freshet Inception Distance (FID) score was used to evaluate the results obtained by this work and as a comparison metric with the literature, where FID score is the most modern metric used to measure the distance between real image and equivalent generated one [54 - 57]. Following are the summary of experiments designed for this step:

- The first experiment validates the work of Lye et al [22] as the model addressing the same problem in literature with the best results.

- In the second experiment, we only replaced the DensePose step in literature by the DensePose proposed in this work.

- The third experiment involves changing only the cloth mask step with the proposed new cloth mask method.

- In the fourth experiment we replaced both the DensePose and cloth mask steps with the new proposed methods.

- In the fifth experiment, we only changed the segmentation model, keeping all the literature model as is.

- The sixth experiment includes deploying the proposed model and evaluating it with all its modules integrated together.

Table II demonstrates results of the six experiments, with "new" referring to the proposed models, and "original" referring to the models in literature [22].

TABLE II. COMPARISON BETWEEN THE STATE-OF-THE-ART APPROACHES AND THE PROPOSED APPROACH

| E# | Segmentation | Cloth Mask | Dense Pose | FID |
|---|---|---|---|---|
| 1 | Original | Original | Original | 11.796 |
| 2 | Original | Original | New | 12.243 |
| 3 | Original | New | Original | 11.847 |
| 4 | Original | New | New | **11.743** |
| 5 | New | Original | Original | 13.140 |
| **6** | **New** | **New** | **New** | **11.753** |

As we can see from the above table the proposed approach outperforms the best model in literature, but it can also be noticed that keeping the segmentation of the literature combined with the proposed Cloth mask and DensePose models would produce even better results, i.e., smaller FID.

### F. User Interface

Finally, to build a complete solution, a user interface has been developed using flutter technology, and the implementation of the application has been divided into two phases. Firstly, creating an initial design by building the main activities, then new features were added using UI/UX. Fig. 16 introduces some of the user interface screens.

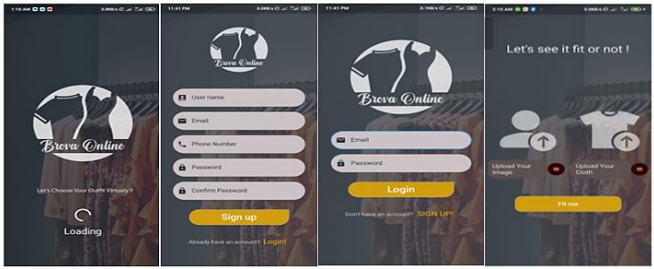

Fig. 16. Some of the final user interface screens.

## VI. Conclusion and Future Work

In conclusion, this work aims at creating a virtual fitting room via a mobile application to make it easier for customers to try on many cloth items without physically dressing them. The best state-of-the-art approach in literature (Lue et al. [22]) has been used in this paper as the baseline architecture to compare the results obtained in this work with. Another reason for choosing them is that they used the Zolando dataset which is publicly available for validation and comparison. But it was found that there is an obvious overfitting because of the unified nature of the dataset. The solution for this included scrapping new data and retraining preprocessing models. A new dataset from different fashion sources was scrapped to collect 5K images to evaluate the proposed approach and compare it to the state-of-the-art approach. All preprocessing subsystems were analyzed and experimented to get the best model. These subsystems included OpenPose (CMU), DensePose (Detectron- R_101_FPN_DL_s1x), cloth mask (cloth segmentation), and segmentation (DeepLab V3). At the end, the proposed approach outperformed the state-of-the-art and succeeded to reduce the FID. The mobile app developed in this work was deployed on AWS service with dockerization technique.

As future work, we suggest adding a new dataset for bottom part and other types of fashion items, in addition to a recommendation system. Other research suggestions may include utilizing newly collected data to retrain the model for improved performance. To reduce inference time semantic segmentation should be considered while training the model, optimizing it for cases with only one instance. Additionally, the use of a teacher-student model (Knowledge Distillation) can be suggested to enhance the performance of all framework models, aiming for better inference times overall. Future work would also include considering children of different ages in testing the proposed architecture.

### Acknowledgment

The authors would like to thank Digital Egypt Builders Initiative (DEBI), at the Egyptian Ministry of Communication and Information Technology (MCIT) for giving us the chance to design and implement this solution, and Amazone Web Services (AWS) whom without their support this work would be never completed.